\documentclass[10pt,twocolumn,letterpaper]{article}

\usepackage{iccv}
\usepackage{times}
\usepackage{epsfig}
\usepackage{graphicx}
\usepackage{amsmath}
\usepackage{amssymb}

\usepackage{caption}
\usepackage{color}

\usepackage{xcolor}

\usepackage{multirow}
\usepackage{colortbl}
\usepackage{bm}
\usepackage{microtype}
\usepackage{amsthm}
\usepackage{subfigure} 
\usepackage{comment}
\usepackage{algorithm}
\usepackage{algpseudocode} 
\usepackage{breqn} 
\usepackage{multirow}
\usepackage{dsfont}
\usepackage{color}

\usepackage[pagebackref=true,breaklinks=true,letterpaper=true,colorlinks,bookmarks=false]{hyperref} 

 \iccvfinalcopy 

\def\iccvPaperID{2445} 

\ificcvfinal\fi

\begin{document}

\title{Generalized Source-free Domain Adaptation}

\author{Shiqi Yang$^{1}$, Yaxing Wang$^{1,2}$\thanks{Corresponding Author.}, Joost van de Weijer$^{1}$, Luis Herranz$^{1}$, Shangling Jui$^{3}$\\
$^{1}$ Computer Vision Center, Universitat Autonoma de Barcelona, Barcelona, Spain\\
$^{2}$ PCALab, Nanjing University of Science and Technology, China\\
$^{3}$ Huawei Kirin Solution, Shanghai, China\\
{\tt\small \{syang,yaxing,joost,lherranz\}@cvc.uab.es}, \tt\small{jui.shangling@huawei.com}
}

\maketitle
\ificcvfinal\thispagestyle{empty}\fi

\begin{abstract}
   Domain adaptation (DA) aims to transfer the knowledge learned from a source domain to an unlabeled target domain. Some recent works tackle source-free domain adaptation (SFDA) where only a source pre-trained model is available for adaptation to the target domain. However, those methods do not consider keeping source performance which is of high practical value in real world applications. In this paper, we propose a new domain adaptation paradigm called Generalized Source-free Domain Adaptation (G-SFDA), where the learned model needs to perform well on both the target and source domains, with only access to current unlabeled target data during adaptation. First, we propose local structure clustering (LSC), aiming to cluster the target features with its semantically similar neighbors, which successfully adapts the model to the target domain in the absence of source data. Second, we propose sparse domain attention (SDA), it produces a binary domain specific attention to activate different feature channels for different domains, meanwhile the domain attention will be utilized to regularize the gradient during adaptation to keep source information.  In the experiments, for target performance our method is on par with or better than existing DA and SFDA methods, specifically it achieves state-of-the-art performance (85.4\%) on VisDA, and our method works well for all domains after adapting to single or multiple target domains. Code is available in \url{https://github.com/Albert0147/G-SFDA}.
\end{abstract}

\section{Introduction}
Though achieving great success, deep neural networks typically require a large amount of labeled data for training. However, collecting labeled data is often laborious and expensive. To tackle this problem, \textit{Domain Adaptation} (DA) methods aim to transfer knowledge learned from label-rich datasets (source domains) to other unlabeled datasets (target domains), by reducing the domain shift between labeled source and unlabeled target domains.

A crucial requirement in most DA methods is that they require access to the source data during adaptation, which is often impossible in many real-world applications, such as deploying domain adaptation algorithms on mobile devices where the computation capacity is limited, or in situations where data-privacy rules limit access to the source domain. Because of its relevance and practical interest, the \textit{source-free domain adaptation} (SFDA) setting, where instead of source data only source pretrained model is available, has started to get traction recently~\cite{kundu2020universal, kundu2020towards, li2020model, liang2020we,  yang2020unsupervised}. Among these methods, SHOT~\cite{liang2020we} and 3C-GAN~\cite{li2020model} are most related to this paper which is for close-set DA where source and target domains have the same categories. 3C-GAN~\cite{li2020model} is based on target-style image generation by a conditional GAN, and SHOT~\cite{liang2020we} proposes to transfer the source hypothesis, i.e. the fixed source classifier, to the target data, together with maximizing mutual information.

However, in many practical situations models should perform well on both the target and source domain. For example, we would desire a recognition model deployed in an urban environment which works well for all four seasons (domains) after adapting model to the seasons sequentially. As shown in~\cite{ye2020light}, the source performance of some DA methods will degrade after adaptation even with source data always at hand. And the current SFDA methods focus on the target domain by fine tuning the source model, leading to forgetting on old domains. Thus, existing methods cannot handle the situation described above.
{A simple way to address this setting is by just storing the source and target model, however, we aim for memory-efficient solutions that scale sub-linear with the number of domains.} Therefore, in this paper, 
we propose a new DA paradigm where the model is expected to perform well on all domains after source-free domain adaptation. We call this setting \textit{Generalized Source-free Domain Adaptation} (G-SFDA). For simplicity, in the paper we will first focus on a single target domain, and then we describe how to extend to Continual Source-free Domain Adaptation.

In this paper, to perform adaptation to the target domain without source data, we first propose Local Structure Clustering (LSC), that clusters each target feature together with its nearest neighbors. The motivation is that one target feature should have similar prediction with its semantic close neighbors. To keep source performance, we propose to use sparse domain attention (SDA), applied to the output of the feature extractor, activating different feature channels depending on the particular domain. The source domain attention will be used to regularize the gradient during target adaptation to prevent forgetting of source information. With LSC and SDA, the adapted model can achieve excellent performance on both source and target domains. In the experiments, we show that for target performance our method is on par with or better than existing DA and SFDA methods on several benchmarks, specifically achieving state-of-the-art performance on VisDA (85.4\%), while simultaneously keeping good source performance. We also extend our method to Continual Source-free Domain Adaptation, where there is more than one target domain, further demonstrating the efficiency of our method.

We summarize our contributions as follows:
\begin{itemize}
\vspace{-1mm}
 
\item We propose a new domain adaptation paradigm denoted as Generalized Source-free Domain Adaptation (G-SFDA), where the source-pretrained model is adapted to target domains while keeping the performance on the source domain, in the absence of source data.

\item We propose local structure clustering (LSC) to achieve source-free domain adaptation, which utilizes local neighbor information in feature space.

\item We propose Sparse domain attention (SDA) which activates different feature channels for different domains, and regularizes the gradient of back propagation during target adaptation to keep information of the source domain.

\item In experiments, we show that where existing methods suffer from forgetting and obtain bad performance on the source domain, our method is able to maintain source domain performance.
Furthermore, when focusing on the target domain our method is on par with or better than existing methods, especially we achieve state-of-the-art target performance on VisDA. 
\end{itemize}

\section{Related Works}
Here we discuss related domain adaptation settings.
\paragraph{Domain Adaptation.}
Early domain adaptation methods such as ~\cite{long2015learning,sun2016return,tzeng2014deep} adopt moment matching to align feature distributions. Inspired by adversarial learning, DANN~\cite{ganin2016domain} formulates domain adaptation as an adversarial two-player game. CDAN~\cite{long2018conditional} trains a deep networks conditioned on several sources of information. DIRT-T~\cite{shu2018dirt} performs domain adversarial training with an added term that penalizes violations of the cluster assumption. Domain adaptation has also been tackled from other perspectives. MCD~\cite{saito2018maximum} adopts prediction diversity between multiple learnable classifiers to achieve local or category-level feature alignment between source and target domains. DAMN~\cite{bermudez2020domain} introduces a framework where each domain undergoes a different sequence of operations. AFN~\cite{Xu_2019_ICCV} shows that the erratic discrimination of target features stems from much smaller norms than those found in source features. SRDC~\cite{tang2020unsupervised} proposes to directly uncover the intrinsic target discrimination via discriminative clustering to achieve adaptation. The most relevant paper to our LSC is DANCE~\cite{saito2020universal}, which is for universal domain adaptation and based on neighborhood clustering. 
But they are based on instance discrimination~\cite{wu2018unsupervised} between all features, while our method applies consistency regularization on only a few semantically close neighbors.

\vspace{-2mm}
\paragraph{Source-free Domain Adaptation.}
Normal domain adaptation methods require access to source data during adaptation. Recently, there are several methods investigating source-free domain adaptation. USFDA~\cite{kundu2020universal} and FS~\cite{kundu2020towards} explore the source-free universal DA~\cite{you2019universal} and open-set DA~\cite{saito2018open}, DECISION~\cite{ahmed2021unsupervised} is for multi-source DA. Related to our work are SHOT~\cite{liang2020we} and 3C-GAN~\cite{li2020model}, both for close-set DA. SHOT proposes to fix the source classifier and match the target features to the fixed classifier by maximizing mutual information and pseudo label. 3C-GAN synthesizes labeled target-style training images based on conditional GAN.  Recently, BAIT~\cite{yang2020unsupervised} extends diverse classifier based domain adaptation methods to also be applicable for SFDA. Though achieving good target performance, these methods cannot maintain source performance after adaptation. Other than these methods, we aim to maintain source-domain performance after adaptation.

\begin{figure*}[tbp]
	\centering
	\includegraphics[width=0.75\textwidth]{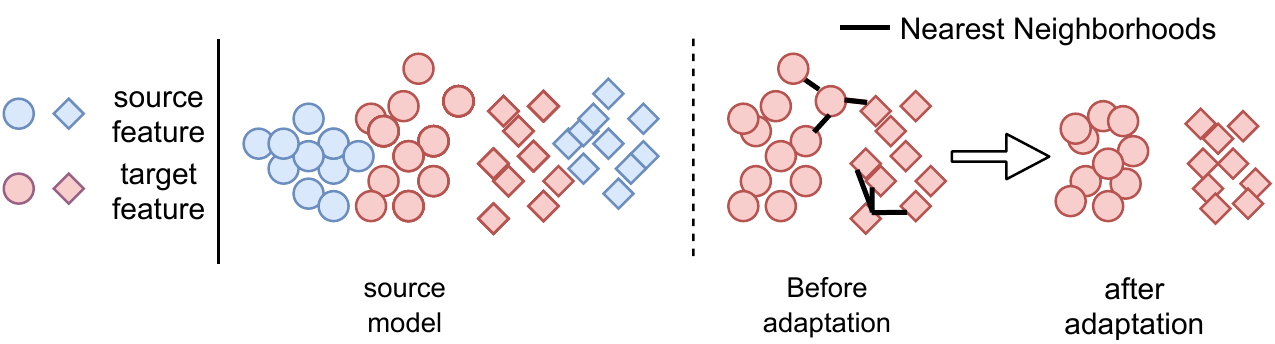}
	\vspace{-2mm}
	\caption{Local Structure Clustering (LSC). Some target features from source model will deviate from dense source feature regions due to domain shift. LSC aims to cluster target features by its semantically close neighbors (linked by black line).\vspace{-0mm}}
	\label{fig:lsc}
	\vspace{-4mm}
\end{figure*}

\vspace{-2mm}
\paragraph{Continual Domain Adaptation.}
Continual learning (CL)~\cite{kirkpatrick2017overcoming,li2017learning,lopez2017gradient,mallya2018packnet} specifically focuses on avoiding catastrophic forgetting when learning new tasks, but it is not tailored for DA since new tasks in CL usually have labeled data. Recently, a few works~\cite{bobu2018adapting,mancini2019adagraph,su2020gradient} have emerged that aim to tackle the \textit{Continual Domain Adaptation} (CDA) problem. \cite{bobu2018adapting} uses sample replay to avoid forgetting together with domain adversarial training, \cite{mancini2019adagraph} builds a domain relation graph, and \cite{su2020gradient} builds a domain-specific memory buffer for each domain to regularize the gradient on both target and memory buffer. Although these methods achieve good performance, they all demand access to source data. And~\cite{kundu2020class} is source-free but they focus on class incremental single target domain adaptation where there is only one-shot labeled target data per class, while our method is related to domain incremental learning and can be deployed for continual source-free domain adaptation.

\section{Methods} 
In this section, we first propose an approach for source-free unsupervised domain adaptation. Then we introduce our method to prevent forgetting of the knowledge of the source model. Next, we elaborate how to unify the two modules to address generalized source-free domain adaptation (G-SFDA), and train a domain classifier for domain-agnostic evaluation. Finally, we extend our method to continual source-free domains.

\subsection{Problem Setting and Notations}
We denote the labeled source domain data with $n_s$, the samples as $\mathcal{D}_s = \{(x_i^s,y^s_i)\}_{i=1}^{n_s}$, where the $y^s_i$ is the corresponding label of $x_i^s$, and the unlabeled target domain data with $n_t$ samples as $\mathcal{D}_t=\{x_j^t\}_{j=1}^{n_t}$. The number of classes is $C$. In the source-free setting we consider here $\mathcal{D}_s$ is only available during model pretraining. Our method is based on a neural network, which we split into two parts: a feature extractor $f$, and a classifier $g$ that only contains one fully connected layer. The output of network is denoted as $p(x)=g(f(x)) \in \mathcal{R}^C$.

\begin{figure*}[t]
	\centering
	\includegraphics[width=0.99\textwidth]{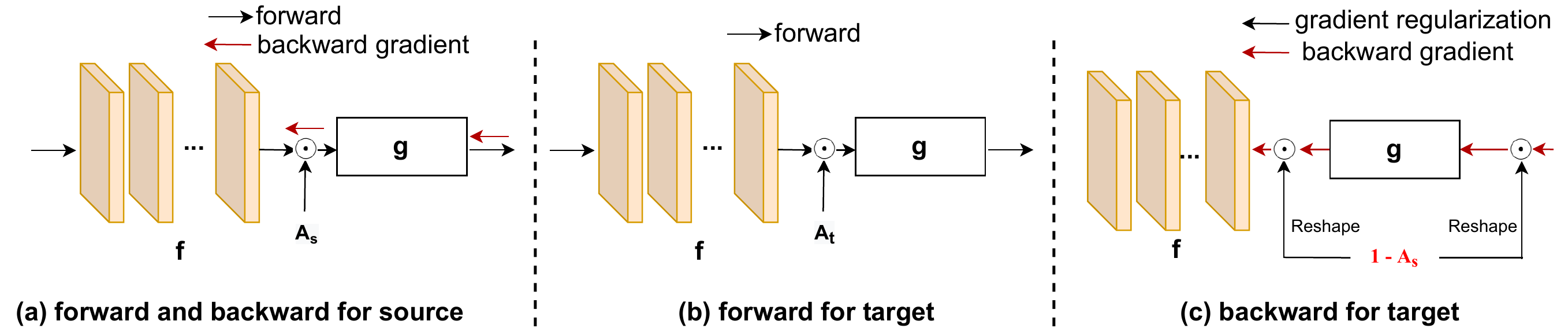}
	\vspace{-3mm}
	\caption{(a-c): Forward and Backward pass for two domains. \textbf{f}, \textbf{g} denote feature extractor, classifier. $\mathcal{A}_s$ and $\mathcal{A}_t$ are the sparse source and target domain attention.\vspace{-2mm}}
	\label{fig:dal}
	\vspace{-2mm}
\end{figure*}

\subsection{Local Structure Clustering}
\label{sec: lsc}
Most domain adaptation methods aim to align the feature distributions of the source and target domain. In source-free unsupervised domain adaptation (SFDA) this is not evident since the algorithm has no longer access to source domain data during adaptation. We identify two main sources of information that the trained source model provides with respect to the target data: a class prediction $p(x)$ and a location in the feature space $f(x)$. The main idea behind our method is that we expect the features of the target domain to be shifted with respect to the source domain, however, we expect that classes still form clusters in the feature space, and as such, we aim to move clusters of data points to their most likely class prediction. 

Our algorithm is illustrated in Fig.~\ref{fig:lsc}~(left). Some target features (at the start of adaptation) deviate from the corresponding dense source feature region due to domain shift. This could result in wrong prediction of the classifier. However, we assume that the target features of the same class are clustered together. Therefore, the nearest neighbors of target features have a high probability to share category labels. To exploit this fact, we encourage features close in feature space to have similar prediction to their nearest neighbors. As a consequences clusters of points that are close in feature space will move jointly towards a common class. As shown in the right of Fig.~\ref{fig:lsc}, this process can correctly classify target features which would otherwise have been wrongly classified. 

To find the semantically close neighbors, we build a feature bank $\mathcal{F}=\{(f(x_i))\}_{x_i \in \mathcal{D}_t}$ which stores the target features. This is similar to methods in unsupervised learning~\cite{wu2018unsupervised,huang2019unsupervised,zhuang2019local,van2020scan} or domain adaptation~\cite{saito2020universal}. The method \cite{saito2020universal} is for universal domain adaptation, and considers similarity based on instance discrimination~\cite{wu2018unsupervised} between all features in their loss function, and~\cite{huang2019unsupervised,van2020scan,zhuang2019local} perform unsupervised learning using neighborhood information. The work \cite{van2020scan} needs pretext training and the nearest neighborhood \textit{images} is retrieved only once by the embedding network from the pretext stage to train another classification network, while \cite{huang2019unsupervised,zhuang2019local} are also based on instance discrimination between all target features, and utilize neighbourhood selection to further improve the cluster performance. Different from them, we only use a few neighbors from the feature bank to cluster the target features with a consistency regularization.

Next, we build
a score bank $\mathcal{S}=\{(g(f(x_i))\}_{x_i \in \mathcal{D}_t}$ storing corresponding softmaxed prediction scores. The local structure clustering is achieved by encouraging consistent predictions between the k-nearest \textit{features} applying the following loss:
\begin{equation}\label{eq:lsc}
\begin{gathered}
\small
\mathcal{L}_{\mathrm{LSC}}=-\frac{1}{n} \sum_{i=1}^{n}\sum_{k=1}^{K}  log[  p(x_i) \cdot s(\mathcal{N}_k)]+\sum_{c=1}^{C} \textrm{KL}(\bar{p}_c||q_c)
\\
\mathcal{N}_{\{1,..,K\}}=\{\mathcal{F}_j | \;top\textrm{-}K\!\! \left(cos\left(f\left(x_i\right),\mathcal{F}_j\right),  \forall \mathcal{F}_j\ \in \mathcal{F}\right) \},\\
 \bar{p}=\frac{1}{n}\sum_{i=1}^n p_{c}(x_i) \ ,\textrm{and}\  q_{\{c=1,..,C\}}= \frac{1}{C} 
\end{gathered}
\end{equation}
Here, we first find the k-nearest neighbors $\mathcal{N}$ in the feature bank for each current target feature based on the cosine similarity.
We minimize the negative log value of the dot product between prediction score of the current target sample $x_i$ and the stored prediction scores $s(\mathcal{N}_k)$ of $\mathcal{N}$, which is the first term in Eq.~\ref{eq:lsc}, aiming to encourage consistent predictions between the feature and its a few neighbors. The second term avoids the degenerated solution~\cite{shi2012information,ghasedi2017deep}, where the prediction of classes in the target data is highly imbalanced, by encouraging prediction balance. Here $p_c$ is the empirical label distribution; it represents the predicted possibility of class $c$ and $q$ is a uniform distribution. And we simply replace the old items in the bank with the new ones corresponding to current mini-batch. In the experiments, we will prove the effectiveness of the proposed LSC by verifying whether the nearest neighbors are sharing the right predicted label.

\subsection{Sparse Domain Attention}

Under the G-SFDA setting, we want to not only have high target performance, but maintain source performance without accessing source data.  Our work is inspired by continual learning (CL) methods~\cite{abati2020conditional,mallya2018packnet,serra2018overcoming} which put constraints on each layer for leaving out capacity for new tasks and prevent forgetting of previous tasks.
We propose to only activate parts of the feature channels of $f(x)\in \mathcal{R}^d$ for different domains, by a sparse domain attention (SDA) vector $\mathcal{A}_{i\in\{s,t\}} \in \mathcal{R}^d$, which contain close-to binary values that will mask the output of the feature extractor. Inspired by \cite{serra2018overcoming}, we adopt an embedding layer to automatically produce the domain adaptation.
\begin{equation}\label{eq:dal}
 \mathcal{A}_{\mathrm{i}\in [s, t]}= \sigma (100 \cdot e_i)
\end{equation}
where $e_i$ is the output of an embedding layer, $\sigma $ is $sigmoid$ function, and the constant 100 is to ensure a near-binary output, but still differentiable. 
$\mathcal{A}_s$ and $\mathcal{A}_t$ are both trained on the source domain and are fixed during the adaptation to the target domain. Furthermore, when training on source, we use sparsity regularization and gradient compensation for the embedding layer just like \cite{serra2018overcoming}.Thus, we use SDA to build domain specific information flows where some channels are specific for each domain. We can maintain the source information by regularizing the gradient flowing into channels that are activated in the source mask. 

For training the source domain, we apply the source attention $\mathcal{A}_s$, as shown in Fig.~\ref{fig:dal}(a), the output is $g(f(x) \odot \mathcal{A}_s)$. 
In Fig.~\ref{fig:dal}(b), we show that when adapting to the target domain, we use the sparse target attention $\mathcal{A}_t$ for the forward pass. To prevent forgetting, there should be no update to the feature channels which are present in $\mathcal{A}_s$. The reasons are twofold: firstly, the information of those channels is the only source information provided during source-free adaptation to the target domain; keeping this information may boost target adaptation, and secondly more importantly, under the G-SFDA setting we hope to keep the source performance after adapting,  therefore target adaptation should not disturb the information flowing to those channels of feature associated with source domain. As shown in Fig.~\ref{fig:dal}(c), during target adaptation we propose to use source attention $A_s$ to regularize the gradients flowing to the classifier and feature extractor during back propagation:
\begin{eqnarray}\label{eq:dal_bpf}
\small
	W_{f_l}\leftarrow W_{f_l} - (\bar{\mathcal A}_s\mathds{1}_{h}^{T}) \odot \frac{\partial \mathcal{L}}{\partial W_{f_l}}\\\label{eq:dal_bpg}
	W_{g}\leftarrow W_{g} -   \frac{\partial \mathcal{L}}{\partial W_{g}} \odot (\mathds{1}_{C}\bar{\mathcal A}_s^T)
\end{eqnarray}
where $\odot$ denotes element wise multiplication, $\mathds{1}_{k}$ is an all-ones vector of dimensionality $k$, $\bar{\mathcal A}_s=1-{\mathcal A}_s$, $W_{f_l} \in \mathcal{R}^{d \times h}$ is the weight of the last layer in feature extractor, $W_{g} \in \mathcal{R}^{C \times d}$ is the weight of the classifier. Here the source attention $\mathcal{A}_s$ is used to regularize the gradient flowing into the source activated channels (for feature extractor) and also the corresponding neurons in the classifier. With Eq.~\ref{eq:dal_bpf} and Eq.~\ref{eq:dal_bpg}, the source information is expected to be preserved.

In continual learning literature the masking of weights~\cite{mallya2018piggyback,mallya2018packnet} and activations~\cite{abati2020conditional,masana2021ternary,serra2018overcoming} has been studied. Our method is related to the activation mask methods. However, other then these methods, our masking only prevents forgetting in the last two layers $W_{f_l}$ and $W_g$.
We ensure that the features that are crucial for source domain performance are only minimally changed, and that the target domain specific features are used to address the domain shift. Our approach does not prevent all forgetting of the source domain, since we do not regularize the gradient of the inner layers in feature extractor.

\subsection{Unified Training}
\begin{algorithm}[tbp]
	\small
	\caption{Generalized Source-free Domain Adaptation}
	\label{alg:ours}
	\begin{algorithmic}[1]
		\Require $\mathcal{D}_s$ (only for source model training), $\mathcal{D}_t$ 
		\State Pre-train model on $\mathcal{D}_s$ with both $\mathcal{A}_s$ and $\mathcal{A}_t$ from SDA
		\State Build feature bank $\mathcal{F}$ and score bank $\mathcal{S}$ for $\mathcal{D}_t$
		\While{Adaptation}
		\State Sample batch $\mathcal{T}$ from $\mathcal{D}_t$ 
		\State Update $\mathcal{F}$ and $\mathcal{S}$ corresponding to current batch $\mathcal{T}$
		\State Compute $\mathcal{L}_{lsc}$ based on $\mathcal{F}$ and $\mathcal{S}$\Comment{Eq.~\ref{eq:lsc},\ref{eq:near}}
		\State Update network with SDA regularization\Comment{Eq.~\ref{eq:dal_bpf},\ref{eq:dal_bpg}}
		\EndWhile 

	\end{algorithmic}
\end{algorithm}
\vspace{-2mm}

In this section, we first illustrate how to unify the training with SDA and LSC. As illustrated in Algorithm~\ref{alg:ours}, first we train the model on $\mathcal{D}_s$ with the cross-entropy loss, with both source and target domain attention $\mathcal{A}_s,\: \mathcal{A}_t$, this is to provide a good initialization for target adaptation where only $\mathcal{A}_t$ is engaged. Then, we adapt the source model to the target domain with target attention $\mathcal{A}_t$ and only access to $\mathcal{D}_t$ with Eq.~\ref{eq:lsc}. During backpropagation we regularize the gradients according to Eq.~\ref{eq:dal_bpf} and Eq.~\ref{eq:dal_bpg}. Unlike training with only LSC in Sec.~\ref{sec: lsc}, here we build the feature bank as $\mathcal{F}=\{(f(x_i)\odot \mathcal{A}_t)\}_{x_i \in \mathcal{D}_t}$, where we abandon the irrelevant channels since those channels will not contribute to current prediction and may contain noise. And for the same reason when using k-nearest neighbors, we also apply the target attention to the feature, so the $\mathcal{N}_{\{1,..,K\}}$ in Eq.~\ref{eq:lsc} turns into:
\begin{equation}\label{eq:near}
\begin{gathered}
\small
\mathcal{N}_{\{1,..,K\}}=\{\mathcal{F}_j | \;top\textrm{-}K\!\! \left(cos\left(f\left(x_i\right)\odot\mathcal{A}_t,\mathcal{F}_j\right),  \forall \mathcal{F}_j\ \in \mathcal{F}\right) \}
\end{gathered}
\end{equation}

\vspace{-2mm}
\paragraph{Domain-ID estimation.} In the experimental section, we will consider both G-SFDA with (\emph{domain-aware}) and without (\emph{domain-agnostic}) access to the domain-id at inference time. In the more challenging setting the domain-ID is not available, and needs to be estimated. 
Therefore, we propose to train a domain classifier which takes in feature $f(x)$ to estimate the domain-ID of the test samples, by only storing a very small set of images of the source domain.  
We will show in the experiments that we obtain similar results in the challenging domain-agnostic setting as in the easier domain-aware setting.

\subsection{Continual Source-free Domain Adaptation}
Here we illustrate how to extend our method to continual source-free domain adaptation, where the model is adapted to a sequence of target domains with only access to current target domain data.
Assuming that there are $N_t$ target domains. For source pretraining we train with all domain attention $\mathcal{A}_s$ and $\{\mathcal{A}_{t_i}\}_{i=1..N_t}$ from SDA, for a good initialization as mentioned before. And when adapting to the $j$-th target domain, we compute $\mathcal{A}^{\prime}$ which considers all domain attention except the current one. We replace the $\mathcal{A}_s$ in Eq.~\ref{eq:dal_bpf} and Eq.~\ref{eq:dal_bpg} with $\mathcal{A}^{\prime}$ for current gradient regularization:
\begin{equation}\label{eq:dal_multi}
\begin{gathered}
\small
 \mathcal{A}^{\prime}= \max (\mathcal{A}^{\prime},\mathcal{A}_{t_i}),\ {\forall i \in \{1,..,N_t\}\setminus j}
\end{gathered}
\end{equation}
where $max$ is an element-wise operation and $\mathcal{A}^{\prime}$ is initialized from $\mathcal{A}_s$. Using $\mathcal{A}^{\prime}$ for gradient regularization means training on one target domain should not influence others.

\begin{table*}[tbp]
	\begin{center}
		\linespread{1.0}
		\scalebox{1.0}{
			\resizebox{\textwidth}{!}{%
			\begin{tabular}{lcccccccccccccc}
			\hline
			Method (Synthesis $\to$ Real)& {Source-free} & plane & bcycl & bus & car & horse & knife & mcycl & person & plant & sktbrd & train & truck & Per-class \\
			\hline
			ResNet-101 \cite{he2016deep}&$\times$  & 55.1 & 53.3 & 61.9 & 59.1 & 80.6 & 17.9 & 79.7 & 31.2  & 81.0 & 26.5  & 73.5 & 8.5  & 52.4   \\
			ADR \cite{saito2017adversarial}&$\times$ & 94.2 & 48.5 & 84.0 & {72.9} & 90.1 & 74.2 &  {92.6} & 72.5 & 80.8 & 61.8 & 82.2 & 28.8 & 73.5 \\
			CDAN \cite{long2018conditional}&$\times$   & 85.2 & 66.9 & 83.0 & 50.8 & 84.2 & 74.9 & 88.1 & 74.5  & 83.4 & 76.0  & 81.9 & 38.0 & 73.9   \\
			CDAN+BSP \cite{chen2019transferability}&$\times$ & 92.4 & 61.0 & 81.0 & 57.5 & 89.0 & 80.6 & 90.1 & 77.0 & 84.2 & 77.9 & 82.1 & 38.4 & 75.9 \\
			SWD \cite{lee2019sliced}&$\times$ & 90.8 & {82.5} & 81.7 & 70.5 & 91.7 & 69.5 & 86.3 & 77.5 & 87.4 & 63.6 & 85.6 & 29.2 & 76.4 \\
			MDD \cite{zhang2019bridging}&$\times$ & - & {-} & -& - & - & - & - & - & - & - & - & - & 74.6 \\
			IA \cite{jiang2020implicit}&$\times$ & - & {-} & -& - & - & - & - & - & - & - & - & - & 75.8 \\
			DMRL \cite{wu2020dual}&$\times$ & - & {-} & -& - & - & - & - & - & - & - & - & - & 75.5 \\
			MCC~\cite{jin2019minimum}&$\times$ & {88.7} & 80.3 & 80.5 & 71.5 & 90.1 & 93.2 & 85.0 & 71.6 & 89.4 & {73.8} & 85.0 & {36.9} & {78.8} \\
			{DANCE}~\cite{saito2020universal}&$\times$ & - & {-} & -& - & - & - & - & - & - & - & - & - & 70.4 \\
			
			\hline
			{DANCE}~\cite{saito2020universal}&\bm{$\surd$} & - & {-} & -& - & - & - & - & - & - & - & - & - & 70.2 \\
			SHOT~\cite{liang2020we}&\bm{$\surd$} &94.3& 88.5 &80.1 &57.3& 93.1&94.9 &80.7&80.3& 91.5& 89.1&86.3& 58.2&\underline{82.9} \\
			3C-GAN~\cite{li2020model}&\bm{$\surd$} &  {94.8} & 73.4 & 68.8 &  {74.8} & 93.1 &  {95.4} & 88.6 &  {84.7} & 89.1 & {84.7} & 83.5 & {48.1} & {81.6} \\
			\hline
		
			 \textbf{Ours w/ domainID}&\bm{$\surd$} & 96.1& 88.3& 85.5& 74.1& 97.1& 95.4 &89.5 &79.4& 95.4 &92.9 &89.1& 42.6 &  \textbf{85.4} \\
			\hline
			\end{tabular}}}
			\caption{Accuracies (\%) on VisDA-C for ResNet101-based unsupervised domain adaptation methods.  {Source-free} means setting without access to source data during adaptation. Underlined results are second highest result. Our results are using target attention $\mathcal{A}_t$. Results of SHOT are from the original paper.}
			\label{tab:visda_da}
		\vspace{-4mm}
	\end{center}
\end{table*}

\begin{table*}[tbb]
	\begin{center}
		\linespread{1.0}
		\addtolength{\tabcolsep}{-4pt}
		\scalebox{0.9}{
			\resizebox{\textwidth}{!}{%
				\begin{tabular}{lcccccccccccccc}
					\hline
					Method& {Source-free}& Ar$\rightarrow$Cl & Ar$\rightarrow$Pr & Ar$\rightarrow$Rw & Cl$\rightarrow$Ar & Cl$\rightarrow$Pr & Cl$\rightarrow$Rw & Pr$\rightarrow$Ar & Pr$\rightarrow$Cl & Pr$\rightarrow$Rw & Rw$\rightarrow$Ar & Rw$\rightarrow$Cl & Rw$\rightarrow$Pr & \textbf{Avg} \\
					\hline\hline
					ResNet-50 \cite{he2016deep} &$\times$& 34.9 & 50.0 & 58.0 & 37.4 & 41.9 & 46.2 & 38.5 & 31.2 & 60.4 & 53.9 & 41.2 & 59.9 & 46.1 \\
					MCD \cite{saito2018maximum} &$\times$&48.9	&68.3	&74.6	&61.3	&67.6	&68.8	&57.0	&47.1	&75.1	&69.1	&52.2	&79.6	&64.1 \\
					CDAN \cite{long2018conditional} &$\times$&50.7	&70.6	&76.0	&57.6	&70.0	&70.0	&57.4	&50.9	&77.3	&70.9	&56.7	&81.6	&65.8 \\
					MDD \cite{zhang2019bridging} &$\times$& 54.9 & 73.7 & 77.8 & 60.0 & 71.4 & 71.8 & 61.2 & {53.6} & 78.1 & 72.5 &  {60.2} & 82.3 & 68.1 \\
					
					IA~\cite{jiang2020implicit} &$\times$& 56.0 & 77.9 & 79.2 & 64.4 & 73.1 & 74.4 & 64.2 & {54.2} & 79.9 & 71.2 & {58.1} & 83.1 & 69.5 \\
					{BNM} \cite{cui2020towards} &$\times$& 52.3 & 73.9 & {80.0} & 63.3 & {72.9} & {74.9} & 61.7 & {49.5} & {79.7} & 70.5 & {53.6} & 82.2 & 67.9 \\
					BDG \cite{yang2020bi} &$\times$& 51.5 & 73.4 & {78.7} & 65.3 & {71.5} & {73.7} & 65.1 & {49.7} & {81.1} & 74.6 & {55.1} & 84.8 & 68.7 \\
					SRDC \cite{tang2020unsupervised} &$\times$& 52.3 & 76.3 & {81.0} &  {69.5} & {76.2} & {78.0} &  {68.7} & {53.8} & {81.7} &  {76.3} & {57.1} &  {85.0} & \underline{71.3} \\
					
					\hline
					SHOT~\cite{liang2020we} &\bm{$\surd$}&57.1 &78.1& 81.5& 68.0& 78.2 &78.1 &67.4 &54.9 &82.2&73.3 &58.8&84.3&\textbf{71.8}  \\
					\hline
				
					\textbf{Ours w/ domainID} &\bm{$\surd$}&57.9&78.6&81.0&66.7&77.2&77.2&65.6&56.0&82.2&72.0&57.8&83.4&\underline{71.3}
 \\
					\hline
		\end{tabular}}}
		\caption{Accuracies (\%) on Office-Home for ResNet50-based unsupervised domain adaptation methods.  {Source-free} means source-free setting without access to source data during adaptation. Underline means the second highest result. Our results are using target attention $\mathcal{A}_t$. Results of SHOT are from the original paper.}
				\label{tab:home_da}
		\vspace{-4mm}
	\end{center}
\end{table*}

\begin{table*}[!tbp]
\centering
\makeatletter 
	\addtolength{\tabcolsep}{-3.5pt}
        \scalebox{1}{
\resizebox{\textwidth}{!}{
\begin{tabular}{l@{~~~~~~~}c cc cc cc cc cc cc ccc}
\hline

       \hline
\hline
  & & \multicolumn{1}{c}{plane } & \multicolumn{1}{c}{bcycl } & \multicolumn{1}{c}{bus }  &  \multicolumn{1}{c}{car} & \multicolumn{1}{c}{horse } & \multicolumn{1}{c}{knife }& \multicolumn{1}{c}{mcycl} & \multicolumn{1}{c}{person} & \multicolumn{1}{c}{plant}  & \multicolumn{1}{c}{sktbrd} & \multicolumn{1}{c}{train} & \multicolumn{1}{c|}{truck}  & \multicolumn{1}{c}{ {\textbf{Avg}.}}& \multicolumn{1}{c}{ } \\

       &    {Source-free}  & \multicolumn{1}{c}{S/T}    & \multicolumn{1}{c}{S/T} &   \multicolumn{1}{c}{S/T}  & \multicolumn{1}{c}{S/T} &   \multicolumn{1}{c}{S/T}  & \multicolumn{1}{c}{S/T}   & S /{T}& S /{T}& S /{T}& S /{T}& S /{T}& \multicolumn{1}{c|}{S /{T}} &    \textbf{S} /{\textbf{T}}&\textbf{H}   \\
\hline
\multicolumn{1}{c}{Source model}   & &  99.9/{70.6}    & 99.9/{15.6} &  99.3/{45.6}   & 99.1/{80.9}   & 99.9/{63.0}  & 99.9/{5.1} & 99.4/{79.2}   & 100/{24.9}    & 99.9/{64.0}    & 100/{39.6}    & 99.3/{84.8}    & \multicolumn{1}{c|}{98.3/{6.3}}   & \textbf{99.6} /{48.1}& 64.9  \\
\hline

\multicolumn{1}{c}{SHOT~\cite{liang2020we}} &\bm{$\surd$}     &  99.3/{94.4}    & 97.3/{85.8} &  34.9/{78.4}   & 47.3/{55.2}   & 94.4/{93.9}  & 93.2/{95.0} & 38.3/{81.5}   & 94.4/{79.5}    & 99.1/{89.8}    & 92.7/{90.1}    & 55.4/{85.6}    &\multicolumn{1}{c|}{62.0/{56.8}}   & 75.7/{82.2}&78.8  \\
\hline

\multicolumn{1}{c}{\textbf{Ours w/ domain-ID}}   &\bm{$\surd$}  & 99.7/{95.9}    & 98.7/{88.1} &  98.4/{85.4}   & 80.0/{72.5}   & 94.6/{96.1}  & 98.4/{93.7}& 76.2/{88.5}   & 97.8/{80.6}    & 98.8/{92.3}    & 99.9/{92.2}    & 75.6/{87.6}    & \multicolumn{1}{c|}{67.3/{44.8}}   & 90.4/{\textbf{85.0}}&\textbf{87.6}    \\
\hline

\multicolumn{1}{c}{\textbf{Ours w/o domain-ID}}   &\bm{$\surd$}  & 99.7/{95.4}    & 98.7/{87.7} &  98.4/{85.7}   & 80.0/{71.5}   & 94.6/{96.1}  & 98.4/{94.8}& 76.2/{89.2}   & 97.8/{80.4}    & 98.8/{92.0}    & 99.9/{88.6}    & 75.6/{87.4}    & \multicolumn{1}{c|}{67.3/{44.1}}   & 90.4/{\underline{84.4}}&\underline{87.3}    \\


\hline
\end{tabular}}
}
\caption{Accuracy (\%) of each method on VisDA dataset using ResNet-101 as backbone under \textbf{G-SFDA} setting. Randomly specifying 0.9/0.1 train/test split for the source dataset. T and S denote accuracy on target and source domain. Domain-ID means having access to domain-ID during evaluation, we provide results under both domain aware and agnostic setting.}
\label{tab:visda_gsfda} 
\end{table*}

\begin{table*}[!tbp]
\centering

\resizebox{\textwidth}{!}{
\begin{tabular}{l@{~~~~~~~}c ccc ccc ccc ccc ccc ccc ccc}
\hline

       \hline
\hline
  & & \multicolumn{3}{c|}{Ar $\rightarrow$ Cl } & \multicolumn{3}{c|}{Ar $\rightarrow$ Pr } & \multicolumn{3}{c|}{Ar $\rightarrow$ Rw }  & 
 
 \multicolumn{3}{c|}{Cl $\rightarrow$ Ar } & \multicolumn{3}{c|}{Cl $\rightarrow$ Pr } & \multicolumn{3}{c}{Cl $\rightarrow$ Rw }  \\
       &    {Source-free}& S & T & \multicolumn{1}{c|}{ {\underline{H}}} &  S & T &  \multicolumn{1}{c|}{ {\underline{H}}} &  S & T &  \multicolumn{1}{c|}{ {\underline{H}}} &  S & T &  \multicolumn{1}{c|}{ {\underline{H}}} &  S & T &  \multicolumn{1}{c|}{ {\underline{H}}} &  S & T &  \multicolumn{1}{c}{ {\underline{H}}} \\
\hline
\multicolumn{1}{c}{Source model}   & &  78.2 & 45.0 & \multicolumn{1}{c|}{57.1}  & 78.2 & 67.2 & \multicolumn{1}{c|}{72.3} & 78.2 & 73.9 & \multicolumn{1}{c|}{76.0} & 79.7 & 49.0 & \multicolumn{1}{c|}{60.7}  & 79.7 & 59.7 & \multicolumn{1}{c|}{68.3}  & 79.7 & 62.2 & \multicolumn{1}{c}{69.9}  \\
\hline

\multicolumn{1}{c}{SHOT~\cite{liang2020we}}   &\bm{$\surd$} &  60.9 & 55.3 & \multicolumn{1}{c|}{58.0}  & 65.2 & 77.4 & \multicolumn{1}{c|}{70.8} & 71.6 & 80.8 & \multicolumn{1}{c|}{75.9} & 65.9 & 68.4 & \multicolumn{1}{c|}{67.1}  & 63.5 & 76.9 & \multicolumn{1}{c|}{69.6}  & 67.4 & 75.7 & \multicolumn{1}{c}{71.3}  \\
\hline
\multicolumn{1}{c}{ \textbf{Ours w/ domain-ID}}   & \bm{$\surd$}&70.0&54.9&\multicolumn{1}{c|}{61.5}&74.0&77.1&\multicolumn{1}{c|}{75.5}&74.5&79.7&\multicolumn{1}{c|}{77.0}&78.5&67.0&\multicolumn{1}{c|}{72.7}&80.3&76.1&\multicolumn{1}{c|}{78.1}&80.6&78.4&\multicolumn{1}{c}{79.5} \\

\hline
\multicolumn{1}{c}{ \textbf{Ours w/o domain-ID}}   & \bm{$\surd$}&68.8&54.7&\multicolumn{1}{c|}{60.9}&72.0&75.6&\multicolumn{1}{c|}{73.8}&74.5&78.5&\multicolumn{1}{c|}{76.4}&77.2&66.6&\multicolumn{1}{c|}{71.5}&79.7&74.0&\multicolumn{1}{c|}{76.7}&78.5&78.4&\multicolumn{1}{c}{78.4} \\

\hline
 & \multicolumn{3}{c|}{Pr $\rightarrow$ Ar } & \multicolumn{3}{c|}{Pr $\rightarrow$ Cl } & \multicolumn{3}{c|}{Pr $\rightarrow$ Rw }
 
  & \multicolumn{3}{c|}{Rw $\rightarrow$ Ar } & \multicolumn{3}{c|}{Rw $\rightarrow$ Cl } & \multicolumn{3}{c|}{Rw $\rightarrow$ Pr }  & \multicolumn{3}{c}{ {\textbf{Avg}.}} \\ 
  
          & S & T & \multicolumn{1}{c|}{ {\underline{H}}} &  S & T &  \multicolumn{1}{c|}{ {\underline{H}}} &  S & T &  \multicolumn{1}{c|}{ {\underline{H}}}&  S & T &  \multicolumn{1}{c|}{ {\underline{H}}}  & S & T &  \multicolumn{1}{c|}{ {\underline{H}}}  & S & T &  \multicolumn{1}{c|}{ {\underline{H}}} & \textbf{S} & \textbf{T} &  \multicolumn{1}{c}{ \textbf{\underline{H}}} \\

\hline
\multicolumn{1}{c}{Source model}  & 92.3 & 52.0 & \multicolumn{1}{c|}{66.5}  & 92.3 & 40.3 & \multicolumn{1}{c|}{56.1}  & 92.3 & 73.0 & \multicolumn{1}{c|}{81.5}  & 85.4 & 64.7 & \multicolumn{1}{c|}{73.6}   & 85.4 & 45.8 & \multicolumn{1}{c|}{59.6}  & 85.4 & 77.5 & \multicolumn{1}{c|}{81.3}  & \textbf{83.9} & 59.2 & \multicolumn{1}{c}{68.6}  \\

\hline
\multicolumn{1}{c}{SHOT~\cite{liang2020we}}  & 78.9 & 65.4 & \multicolumn{1}{c|}{71.5}  & 74.2 & 54.2 & \multicolumn{1}{c|}{62.6}  & 84.9 & 80.5 & \multicolumn{1}{c|}{82.6}  & 79.7 & 71.7 & \multicolumn{1}{c|}{75.5}   & 71.0 & 59.0 & \multicolumn{1}{c|}{64.4}  & 79.2 & 84.6 & \multicolumn{1}{c|}{81.8}  & 71.9 & \textbf{70.8} & \multicolumn{1}{c}{70.9}  \\
\hline

\multicolumn{1}{c}{ \textbf{Ours w/ domain-ID}}  &89.8&65.7&\multicolumn{1}{c|}{75.9}&89.3&53.8&\multicolumn{1}{c|}{67.1}&91.6&81.9&\multicolumn{1}{c|}{86.5}&85.9&71.5&\multicolumn{1}{c|}{78.0}&81.3&60.5&\multicolumn{1}{c|}{69.4}&84.4&83.4&\multicolumn{1}{c|}{83.9}&\underline{81.8}&\textbf{70.8} & \multicolumn{1}{c}{ {$\textbf{75.5}$}}\\
\hline

\multicolumn{1}{c}{ \textbf{Ours w/o domain-ID}}  &87.8&65.1&\multicolumn{1}{c|}{74.8}&86.3&53.2&\multicolumn{1}{c|}{65.8}&90.3&81.6&\multicolumn{1}{c|}{85.7}&83.2&72.0&\multicolumn{1}{c|}{77.2}&78.3&60.2&\multicolumn{1}{c|}{68.1}&83.4&82.8&\multicolumn{1}{c|}{83.1}&80.0&\underline{70.2} & \multicolumn{1}{c}{ {$\underline{74.4}$}}\\
\hline

\hline
\end{tabular}
}
\caption{Accuracy (\%) of each method on Office-Home dataset using ResNet-50 as backbone under \textbf{G-SFDA} setting. Randomly specifying 0.8/0.2 train/test split for the source dataset. T and S denote accuracy on target and source domain. domain-ID means having access to domain-ID during evaluation, w/o domain-ID means using the estimated domain-ID from domain classifier.}
\label{tab:home_gsfda} 
\end{table*}

\begin{figure}[!tbp]
	\centering
	\includegraphics[width=0.49\textwidth]{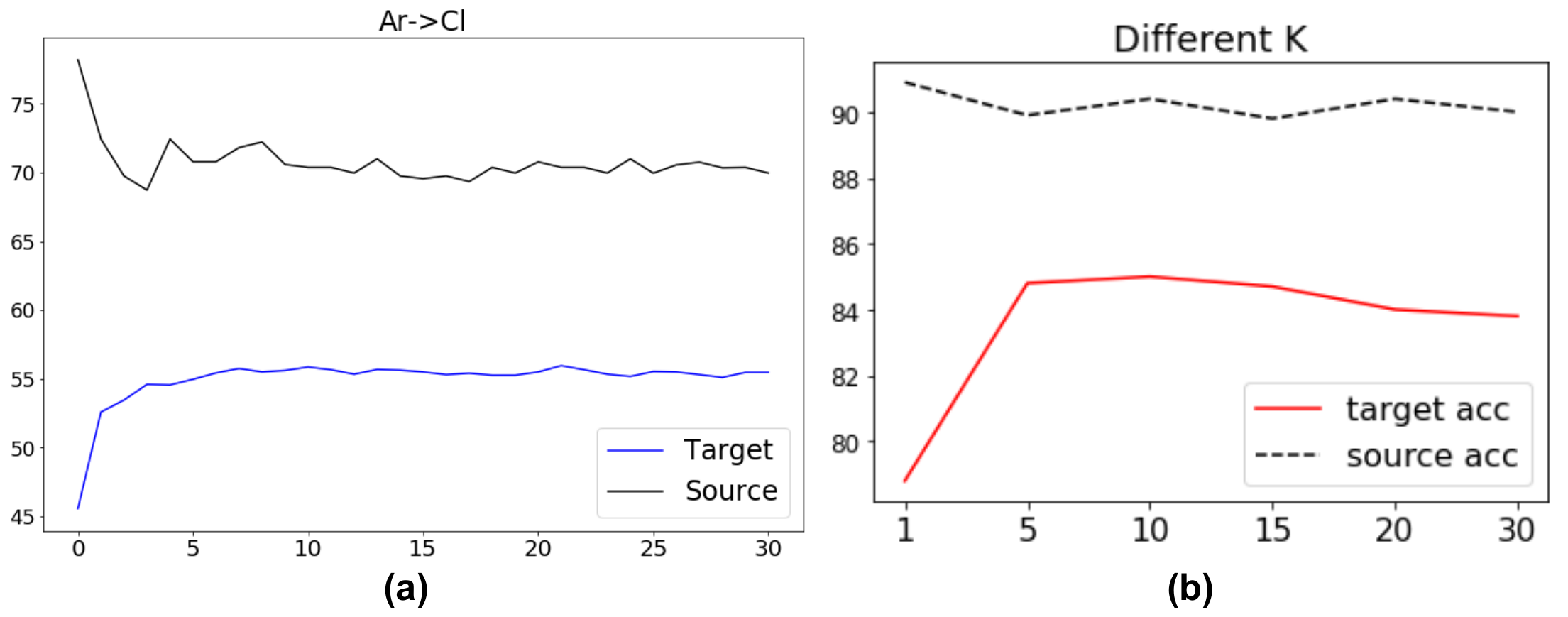}
	\vspace{-6mm}
	\caption{(a) Training curves on task Ar$\rightarrow$Cl of Office-Home dataset. (b) Ablation study of different $K$ on VisDA.\vspace{-0mm}}
	\label{fig:curves_k}
	\vspace{-2mm}
\end{figure}

\begin{table*}[!tbp]
 \centering
  \centering
     \makeatletter 
        \scalebox{0.8}{
       \begin{tabular}{ccc}
					\hline
					$\textbf{Office-Home}$&S&T\\
					\hline
				Source model &{\textbf{83.9}}&59.2\\
	          \hline
				Ours (w/o SDA) &{72.4}&70.2\\
				Ours (w/ SDA)&{81.8}&\textbf{70.8}\\
					\hline
		\end{tabular}}
\scalebox{0.8}{\begin{tabular}{ccc}
					\hline
					$\textbf{VisDA}$&S&T\\
					\hline
				Source model &{\textbf{99.6}}&48.1\\
	          \hline
				Ours (w/o SDA) &{72.1}&74.6\\
				Ours (w/ SDA)&{90.4}&\textbf{85.0}\\
					\hline
		\end{tabular}}
  \centering
     \makeatletter 
     \addtolength{\tabcolsep}{-4pt}
        \scalebox{0.8}{
       \begin{tabular}{ccc}
					\hline
					$\textbf{OH /$s$}$&S&T\\
					\hline
			65 (paper)  &{{80.0}}&70.2\\
				130  &{80.6}&70.3\\
				195 &\textbf{80.8}&\textbf{70.4}\\
					\hline
		\end{tabular}}
  \centering
     \makeatletter 
     \addtolength{\tabcolsep}{2pt}
\scalebox{0.8}{\begin{tabular}{ccc}
					\hline
					$\textbf{VisDA \text/$s$}$&S&T\\
					\hline
				16 &{89.0}&{83.6}\\
				32 &{90.2}&84.2\\
			64 (paper)  &{\textbf{90.4}}&\textbf{84.4}\\
					\hline
		\end{tabular}}
		

   \vspace{-2mm}
   \caption{(\textbf{Left} two) Ablation study on Office-Home and VisDA. The S and T means source and target accuracy. (\textbf{Right} two) Ablation on number of stored images per domain to train domain classifier. \vspace{-2mm}}\label{tab:aba_dc}
\end{table*}

\begin{table*}[!tbp]
 \begin{minipage}[tbp]{0.24\textwidth}
     \makeatletter
       \scalebox{0.9}{
			\resizebox{\textwidth}{!}{\begin{tabular}{|c|c|c|c|c|} 
\hline
\multirow{2}{*}{} & \multicolumn{4}{c|}{test}  \\ 
\cline{2-5}
                  & Ar   & Cl   & Pr   & Rw    \\ 
\hline
Ar                & 74.5 & 42.0 & 61.3 & 68.2  \\ 
\hline
Cl                & 71.4 & 56.6 & 61.2 & 67.9  \\ 
\hline
Pr                & 70.9 & 55.7 & 73.0 & 71.2  \\ 
\hline
Rw                & 72.6 & 55.6 & 72.7 & 77.2  \\
\hline
\end{tabular}}}
  \end{minipage}
  \begin{minipage}[tbp]{0.24\textwidth}
        \makeatletter 
        \scalebox{0.9}{
			\resizebox{\textwidth}{!}{\begin{tabular}{|c|c|c|c|c|} 
\hline
\multirow{2}{*}{} & \multicolumn{4}{c|}{test}  \\ 
\cline{2-5}
                  & Cl   & Ar   & Pr   & Rw    \\ 
\hline
Cl                & 82.2 & 49.7 & 60.0 & 61.2  \\ 
\hline
Ar                & 80.1 & 65.4 & 63.7 & 66.3  \\ 
\hline
Pr                & 79.7 & 63.2 & 72.9 & 68.2  \\ 
\hline
Rw                & 78.6 & 64.9 & 72.8 & 72.4  \\
\hline
\end{tabular}}}
   \end{minipage}
   \begin{minipage}[tbp]{0.24\textwidth}
        \makeatletter 
        \scalebox{0.9}{
			\resizebox{\textwidth}{!}{\begin{tabular}{|c|c|c|c|c|} 
\hline
\multirow{2}{*}{} & \multicolumn{4}{c|}{test}  \\ 
\cline{2-5}
                  & Pr   & Ar   & Cl   & Rw    \\ 
\hline
Pr                & 92.0 & 49.7 & 41.0 & 71.0  \\ 
\hline
Ar                & 91.0 & 63.6 & 42.7 & 72.6  \\ 
\hline
Cl                & 89.2 & 61.8 & 53.1 & 70.4  \\ 
\hline
Rw                & 88.6 & 63.1 & 51.5 & 76.5  \\
\hline
\end{tabular}}}
   \end{minipage}
   \begin{minipage}[tbp]{0.24\textwidth}
        \makeatletter 
        \scalebox{0.9}{
			\resizebox{\textwidth}{!}{\begin{tabular}{|c|c|c|c|c|} 
\hline
\multirow{2}{*}{} & \multicolumn{4}{c|}{test}  \\ 
\cline{2-5}
                  & Rw   & Ar   & Cl   & Pr    \\ 
\hline
Rw                & 86.0 & 63.0 & 45.7 & 77.6  \\ 
\hline
Ar                & 85.7 & 72.4 & 49.8 & 77.4  \\ 
\hline
Cl                & 80.7 & 68.9 & 59.1 & 73.4  \\ 
\hline
Pr                & 84.2 & 69.1 & 57.4 & 80.5  \\
\hline
\end{tabular}}}
   \end{minipage}
   \vspace{-2mm}
   \caption{Continual Source-free Domain Adaptation, the model is adapted from source domain (the first domain) to all target domain sequentially. The results on source domain are reported on the test set.\vspace{-4mm}}\label{tab:multi}\vspace{-2mm}
\end{table*}

\begin{figure}[!tbp]
	\centering
	\includegraphics[width=0.49\textwidth]{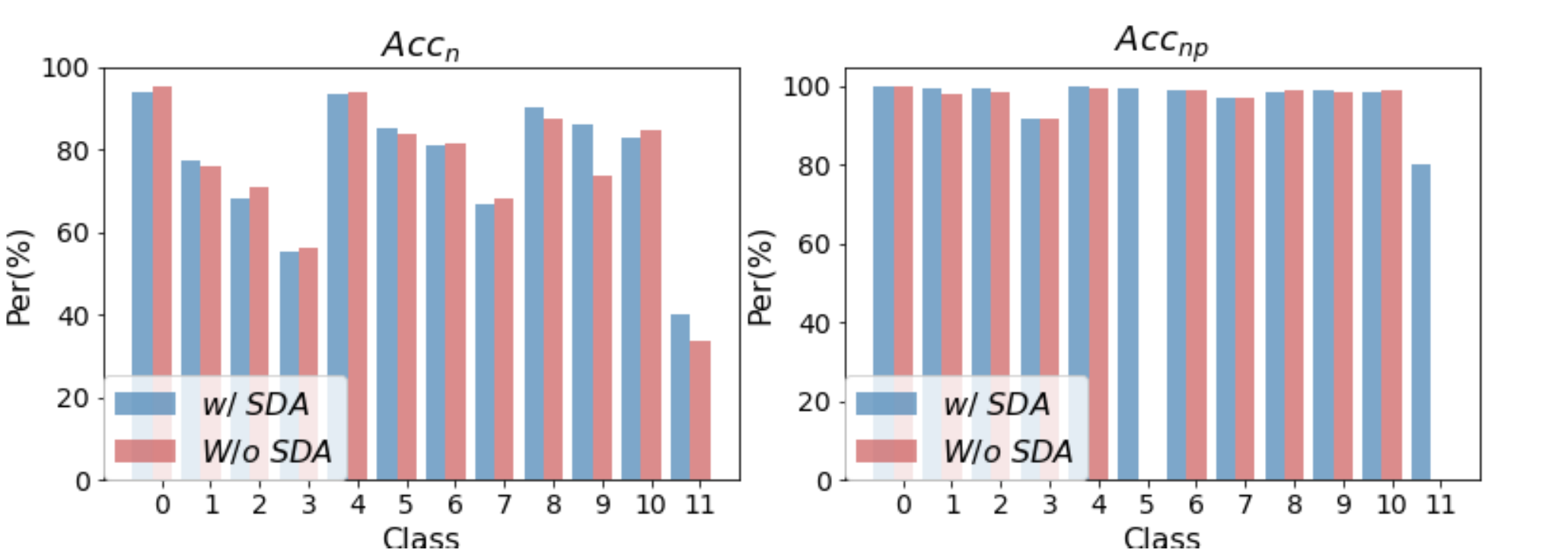}
	\vspace{-8mm}
	\caption{Ablation study of SDA on VisDA, which has 12 classes. $Acc_n$ means the percentage of target features which share the same \textbf{predicted} label with its 3 nearest neighbors, and $Acc_{np}$ means the percentage among above features which have the \textbf{correct} shared {predicted} class.\vspace{-2mm}}
	\label{fig:neighbor}
	\vspace{-2mm}
\end{figure}

\section{Experiments}
\noindent \textbf{Datasets.} \emph{Office-Home}~\cite{venkateswara2017deep} contains 4 domains (Real, Clipart, Art, Product) with 65 classes and a total of 15,500 images. \emph{VisDA}~\cite{peng2017visda} is a more challenging dataset with 12 classes. Its source domain contains 152k synthetic images while the target domain has 55k real object images. 

\noindent \textbf{Evaluation.} We mainly compare with existing methods under two different settings, one is the normal DA and SFDA setting where target performance is the only focus. Another is our proposed G-SFDA setting, where the adapted model is expected to have good performance on both source and target domains after source-free domain adaptation. In this setting, we compute the harmonic mean between source and target accuracy: $H=\frac{2*Acc_S*Acc_T}{Acc_S+Acc_T}$, and $Acc_S$ and $Acc_T$ are respectively the accuracy on source and target test data. For SFDA, we use all source data for model pretraining. And for G-SFDA we only use part (80\% for Office-Home and 90\% for VisDA), the remaining source data is used for evaluating source performance. We provide results under both the domain aware and domain agnostic setting (where we estimate the domain-ID with the domain classifier). Finally, we report results for continual source-free domain adaptation.

\noindent \textbf{Model details.} We adopt the backbone of ResNet-50~\cite{he2016deep} for Office-Home and  ResNet-101 for VisDA along with an extra fully connected (fc) layer as feature extractor, and a fc layer as classifier head. We adopt SGD with momentum 0.9 and batch size of 64 on all datasets. The learning rate for Office-Home is set to 1e-3 for all layers, except for the last two newly added fc layers, where we apply 1e-2. Learning rates are set 10 times smaller for VisDA. \textit{On the source domain, we train the whole network with all domain attentions from SDA, while for target adaptation, we only train the BN layers and last layer in feature extractor, as well as the classifier}. We train 30 epochs on the target domain for Office-Home while 15 epochs for VisDA. For the number of nearest neighbors ($K$) in Eq.~\ref{eq:lsc}, we use 2 for Office-Home, since VisDA is much larger we set $K$ to 10. All results are the average between three runs with random seeds. For training the domain classifier, we store one image per class for Office-Home (total 130 images for 65 classes, 2 domains), and randomly sample 64 images per domain for VisDA (total 128 images for 12 classes, 2 domains). The domain classifier only contains 2 fc layers. 

\subsection{Comparing with State-of-the-art}
\paragraph{Target-oriented Domain Adaptation.}
We first evaluate the target performance of our method compared with existing DA and SFDA methods. The results on the VisDA and Office-Home dataset are shown in Tab.~\ref{tab:visda_da}-\ref{tab:home_da}, our results are using target attention $\mathcal{A}_t$. In these tables, the top part (denoted by $\times$ in the \textit{source-free} column) shows results for the normal setting with access to source data during adaptation. The bottom one (denoted by $\surd$ in the \textit{source-free} column) shows results for the source-free setting. 
Our method achieves state-of-the-art performance on VisDA surpassing SHOT by a large margin (2.5\%). 
The reported results clearly demonstrate the efficiency of the proposed method for source-free domain adaptation. Interestingly, like already observed in the SHOT paper, source-free methods outperform methods that have access to source data during adaptation. Our method is on par with existing DA methods on Office-Home, where our method gets the same results as the DA method SRDC~\cite{tang2020unsupervised} and is a little inferior to the SFDA method SHOT (0.5\% lower than SHOT).
{In addition, we show the results of DANCE~\cite{saito2020universal} with and without source data in Tab.~\ref{tab:visda_da} which are almost the same. Since both of DANCE and our method are using neighborhood information for adaptation, these results may imply that source data are not necessity when efficiently exploiting the target feature structure.}

\vspace{-2mm}
\paragraph{Generalized Source-free Domain Adaptation.}
Here we evaluate our method under the G-SFDA setting. Since we leave out part of the source data for evaluation, we need to reproduce current SFDA methods. 3C-GAN~\cite{li2020model} did not release code, we therefore only compare with the source-free method SHOT~\cite{liang2020we} reproduced by ourselves based on the author's code.
As shown in Tab.~\ref{tab:visda_gsfda}-\ref{tab:home_gsfda}, first our method (w/ domain-ID) obtains a significantly higher $H$ value improving SHOT by 8.8\% on Office-Home and 4.6\% on VisDA. The gain is mainly due to superior results on the source dataset, since SHOT suffers from forgetting. Compared with the source model, our method still has a drop of 2.1\% and 9.2\% lower on Office-Home and VisDA, implying there is still space to explore further techniques to reduce forgetting. 
We also report the results for domain agnostic evaluation, where we use the domain classifier to estimate domain-ID. As shown in the last row of Tab.~\ref{tab:visda_gsfda} and Tab.~\ref{tab:home_gsfda}, with the estimated domain-ID, our methods can get similar results compared with the domain aware method, and still report superior $H$ values compared to SHOT.
{Note there is still source performance degradation, since we only deploy one SDA module before the classifier. The forgetting is caused in the layers inside the feature extractor.
One factor is the statistics in the BN layers which will be replaced by the target statistics after adaptation. If we would adapt the BN parameters back to the source domain (by simply doing a forward pass to update BN statistics before evaluation), we found that this leads to a performance gain (0.7\% and 1.6\% on Office-Home and VisDA respectively) on the source domain.}

\subsection{Analysis and further experiments}

\paragraph{Training curves.} As shown in Fig.~\ref{fig:curves_k}(a), with SDA the source performance during the whole adaptation stage is quite smooth, which proves the efficiency of SDA.

\vspace{-2mm}
\paragraph{Number of nearest neighbors $K$.} In Fig.~\ref{fig:curves_k}(b), we show the results with different $K \in \{1,5,10,15,20,30\}$ in Eq.~\ref{eq:lsc} on VisDA. Our method is quite robust to the choice of $K$, only $K$ is 1 results in lower results. We conjecture that only using a single nearest neighbor in Eq.\ref{eq:lsc} maybe noisy if the feature locates in dense regions.  

\vspace{-2mm}
\paragraph{Ablation study of SDA.} We show the results of removing the SDA in the left of Tab.~\ref{tab:aba_dc}. As expected removing SDA leads to a large drop in source performance. Unexpected is that removing SDA also deteriorates target performance: a lot on VisDA (10.4$\downarrow$), and a little for Office-Home (0.6$\downarrow$). To further investigate it, we check how well LSC works with and without SDA on VisDA in Fig.~\ref{fig:neighbor}; here $Acc_n$ means the percentage of target features which share the same predicted label with its 3 nearest neighbors, and among those features $Acc_{np}$ means the percentage having the correct shared predicted label. According to the results, LSC can lead to good local structure (most neighbors share the same prediction), however the prediction maybe wrong if removing SDA, this is especially the case for class 5 and 11 which have totally wrong prediction ($Acc_{np}$ is 0). This may imply keeping source information with SDA is helping target adaptation.

\vspace{-2mm}
\paragraph{Domain classifier.} 
{We report results as a function of the number of stored images for training domain classifier (right of Tab.~\ref{tab:aba_dc}). For Office-Home, we ensure at least one image per class. The results show with a small amount of stored images, the learned domain-ID classifier works well.
}

\begin{figure}[!tbp]
	\centering
	\includegraphics[width=0.48\textwidth]{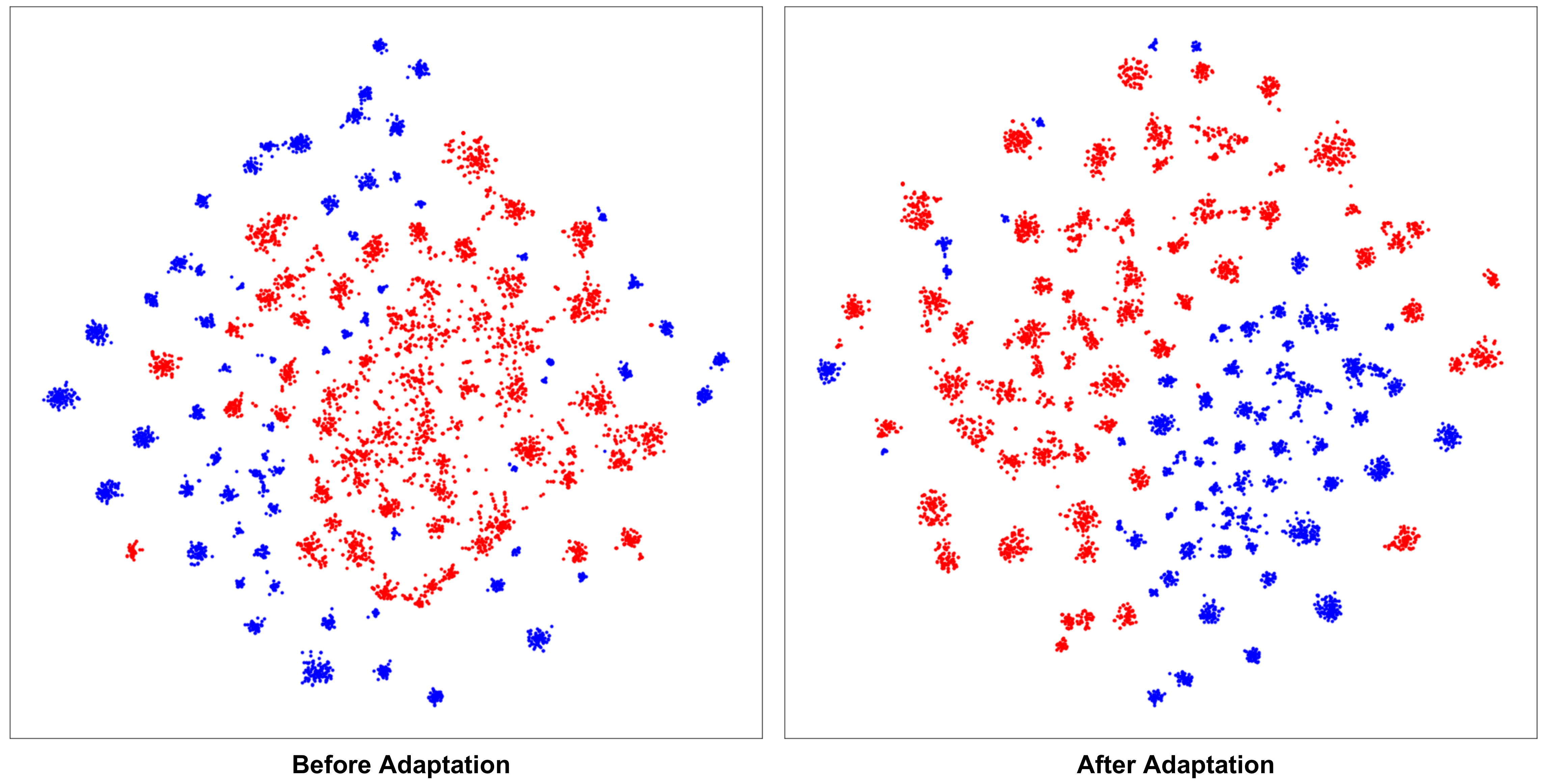}\vspace{-3mm}
	\caption{t-SNE visualization of features before and after adaptation on task Ar$\rightarrow$Pr of Office-Home. The blue are source features while the red are target.\vspace{-2mm}}
	\label{fig:tsne}
	\vspace{-2mm}
\end{figure}

\begin{figure}[!tbp]
	\centering
	\includegraphics[width=0.48\textwidth]{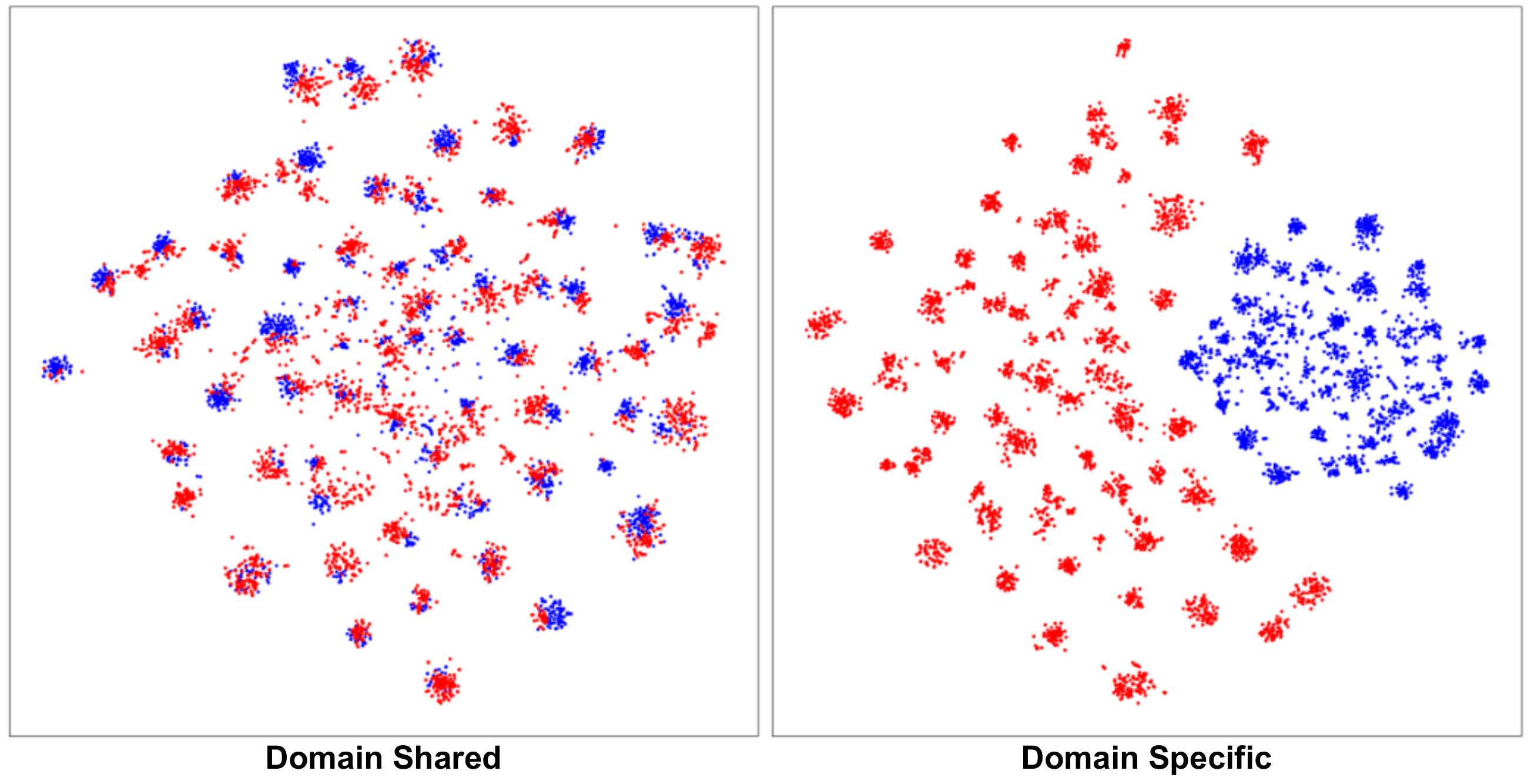}\vspace{-3mm}
	\caption{t-SNE of features from domain shared and domain specific channels after adaptation (task Ar$\rightarrow$Pr on Office-Home). The blue are source features while red for target.\vspace{-2mm}}
	\label{fig:tsne_ss}
	\vspace{-6mm}
\end{figure}

\vspace{-2mm}
\paragraph{t-SNE visualization.} We visualize the features before and after adaptation, which are already masked by the different domain attentions, the source and target features are expected to cluster independently, just as shown in Fig.~\ref{fig:tsne}. The source clusters maintain well after adaptation, and the disordered target features turn into more structured after adaptation. We also visualize features in the shared and specific domain channels. As shown in Fig.~\ref{fig:tsne_ss}, features in the shared domain channels cluster together, but features in the specific domain channels are totally separated across domains.

\vspace{-2mm}
\paragraph{Continual Source-free Domain Adaptation.}
We also provide results (domain aware) of continual source-free domain adaptation in Tab.~\ref{tab:multi}. The results show that it can work well for all domains. The interesting thing is that adapting to one target domain will improve the performance on not-seen target domain, for example, when adapting the model from source domain \emph{Cl} to the first target domain \emph{Ar}, the unseen target domain \emph{Rw} also gains. The reason is that the information learned currently is also helpful for future target domain. Note for some target domains, the result is lower compared with directly adapting from source to the domain, the reason is that we decrease the learned channels by using more gradient regularization as in Eq.~\ref{eq:dal_multi}, implying more capacity is needed for adapting to more domains.

\vspace{-2mm}
\section{Conclusion}
In this paper, we propose a new domain adaptation paradigm denoted as Generalized Source-free Domain Adaptation, where the learned model needs to have good performance on both the target and source domains, with only access to the unlabeled target domain during adaptation. We propose local structure clustering to keep local target cluster information in feature space, successfully adapting the model to the target domain without source domain data. We propose sparse domain attention, which activates different feature channels for different domains, and is also utilized to regularize the gradient during target training to maintain source domain information. 
Experiment results testify the efficacy of our method.

\small{\textbf{Acknowledgement} We acknowledge the support from Huawei Kirin Solution, and the project PID2019-104174GB-I00 (MINECO, Spain) and RTI2018-102285-A-I00 (MICINN, Spain), Ramón y Cajal fellowship RYC2019-027020-I, and the CERCA Programme of Generalitat de Catalunya.}

{\small
\bibliographystyle{ieee_fullname}
\bibliography{egbib}

\begin{thebibliography}{10}\itemsep=-1pt

\bibitem{abati2020conditional}
Davide Abati, Jakub Tomczak, Tijmen Blankevoort, Simone Calderara, Rita
  Cucchiara, and Babak~Ehteshami Bejnordi.
\newblock Conditional channel gated networks for task-aware continual learning.
\newblock In {\em Proceedings of the IEEE/CVF Conference on Computer Vision and
  Pattern Recognition}, pages 3931--3940, 2020.

\bibitem{ahmed2021unsupervised}
Sk~Miraj Ahmed, Dripta~S Raychaudhuri, Sujoy Paul, Samet Oymak, and Amit~K
  Roy-Chowdhury.
\newblock Unsupervised multi-source domain adaptation without access to source
  data.
\newblock In {\em Proceedings of the IEEE/CVF Conference on Computer Vision and
  Pattern Recognition}, pages 10103--10112, 2021.

\bibitem{bermudez2020domain}
Roger Bermudez~Chacon, Mathieu Salzmann, and Pascal Fua.
\newblock Domain-adaptive multibranch networks.
\newblock In {\em 8th International Conference on Learning Representations},
  2020.

\bibitem{bobu2018adapting}
Andreea Bobu, Eric Tzeng, Judy Hoffman, and Trevor Darrell.
\newblock Adapting to continuously shifting domains.
\newblock 2018.

\bibitem{chen2019transferability}
Xinyang Chen, Sinan Wang, Mingsheng Long, and Jianmin Wang.
\newblock Transferability vs. discriminability: Batch spectral penalization for
  adversarial domain adaptation.
\newblock In {\em International Conference on Machine Learning}, pages
  1081--1090, 2019.

\bibitem{cui2020towards}
Shuhao Cui, Shuhui Wang, Junbao Zhuo, Liang Li, Qingming Huang, and Qi Tian.
\newblock Towards discriminability and diversity: Batch nuclear-norm
  maximization under label insufficient situations.
\newblock {\em CVPR}, 2020.

\bibitem{ganin2016domain}
Yaroslav Ganin, Evgeniya Ustinova, Hana Ajakan, Pascal Germain, Hugo
  Larochelle, Fran{\c{c}}ois Laviolette, Mario Marchand, and Victor Lempitsky.
\newblock Domain-adversarial training of neural networks.
\newblock {\em The Journal of Machine Learning Research}, 17(1):2096--2030,
  2016.

\bibitem{ghasedi2017deep}
Kamran Ghasedi~Dizaji, Amirhossein Herandi, Cheng Deng, Weidong Cai, and Heng
  Huang.
\newblock Deep clustering via joint convolutional autoencoder embedding and
  relative entropy minimization.
\newblock In {\em Proceedings of the IEEE international conference on computer
  vision}, pages 5736--5745, 2017.

\bibitem{he2016deep}
Kaiming He, Xiangyu Zhang, Shaoqing Ren, and Jian Sun.
\newblock Deep residual learning for image recognition.
\newblock In {\em Proceedings of the IEEE conference on computer vision and
  pattern recognition}, pages 770--778, 2016.

\bibitem{huang2019unsupervised}
Jiabo Huang, Qi Dong, Shaogang Gong, and Xiatian Zhu.
\newblock Unsupervised deep learning by neighbourhood discovery.
\newblock In {\em International Conference on Machine Learning}, pages
  2849--2858. PMLR, 2019.

\bibitem{jiang2020implicit}
Xiang Jiang, Qicheng Lao, Stan Matwin, and Mohammad Havaei.
\newblock Implicit class-conditioned domain alignment for unsupervised domain
  adaptation.
\newblock {\em arXiv preprint arXiv:2006.04996}, 2020.

\bibitem{jin2019minimum}
Ying Jin, Ximei Wang, Mingsheng Long, and Jianmin Wang.
\newblock Minimum class confusion for versatile domain adaptation.
\newblock {\em ECCV}, 2020.

\bibitem{kirkpatrick2017overcoming}
James Kirkpatrick, Razvan Pascanu, Neil Rabinowitz, Joel Veness, Guillaume
  Desjardins, Andrei~A Rusu, Kieran Milan, John Quan, Tiago Ramalho, Agnieszka
  Grabska-Barwinska, et~al.
\newblock Overcoming catastrophic forgetting in neural networks.
\newblock {\em Proceedings of the national academy of sciences},
  114(13):3521--3526, 2017.

\bibitem{kundu2020universal}
Jogendra~Nath Kundu, Naveen Venkat, and R~Venkatesh Babu.
\newblock Universal source-free domain adaptation.
\newblock {\em CVPR}, 2020.

\bibitem{kundu2020towards}
Jogendra~Nath Kundu, Naveen Venkat, Ambareesh Revanur, R~Venkatesh Babu, et~al.
\newblock Towards inheritable models for open-set domain adaptation.
\newblock In {\em Proceedings of the IEEE/CVF Conference on Computer Vision and
  Pattern Recognition}, pages 12376--12385, 2020.

\bibitem{kundu2020class}
Jogendra~Nath Kundu, Rahul~Mysore Venkatesh, Naveen Venkat, Ambareesh Revanur,
  and R~Venkatesh Babu.
\newblock Class-incremental domain adaptation.
\newblock {\em ECCV}, 2020.

\bibitem{lee2019sliced}
Chen-Yu Lee, Tanmay Batra, Mohammad~Haris Baig, and Daniel Ulbricht.
\newblock Sliced wasserstein discrepancy for unsupervised domain adaptation.
\newblock In {\em Proceedings of the IEEE Conference on Computer Vision and
  Pattern Recognition}, pages 10285--10295, 2019.

\bibitem{li2020model}
Rui Li, Qianfen Jiao, Wenming Cao, Hau-San Wong, and Si Wu.
\newblock Model adaptation: Unsupervised domain adaptation without source data.
\newblock In {\em Proceedings of the IEEE/CVF Conference on Computer Vision and
  Pattern Recognition}, pages 9641--9650, 2020.

\bibitem{li2017learning}
Zhizhong Li and Derek Hoiem.
\newblock Learning without forgetting.
\newblock {\em IEEE transactions on pattern analysis and machine intelligence},
  40(12):2935--2947, 2017.

\bibitem{liang2020we}
Jian Liang, Dapeng Hu, and Jiashi Feng.
\newblock Do we really need to access the source data? source hypothesis
  transfer for unsupervised domain adaptation.
\newblock {\em ICML}, 2020.

\bibitem{long2015learning}
Mingsheng Long, Yue Cao, Jianmin Wang, and Michael~I Jordan.
\newblock Learning transferable features with deep adaptation networks.
\newblock {\em ICML}, 2015.

\bibitem{long2018conditional}
Mingsheng Long, Zhangjie Cao, Jianmin Wang, and Michael~I Jordan.
\newblock Conditional adversarial domain adaptation.
\newblock In {\em Advances in Neural Information Processing Systems}, pages
  1647--1657, 2018.

\bibitem{lopez2017gradient}
David Lopez-Paz and Marc'Aurelio Ranzato.
\newblock Gradient episodic memory for continual learning.
\newblock In {\em Proceedings of the 31st International Conference on Neural
  Information Processing Systems}, pages 6470--6479, 2017.

\bibitem{mallya2018piggyback}
Arun Mallya, Dillon Davis, and Svetlana Lazebnik.
\newblock Piggyback: Adapting a single network to multiple tasks by learning to
  mask weights.
\newblock In {\em Proceedings of the European Conference on Computer Vision
  (ECCV)}, pages 67--82, 2018.

\bibitem{mallya2018packnet}
Arun Mallya and Svetlana Lazebnik.
\newblock Packnet: Adding multiple tasks to a single network by iterative
  pruning.
\newblock In {\em Proceedings of the IEEE Conference on Computer Vision and
  Pattern Recognition}, pages 7765--7773, 2018.

\bibitem{mancini2019adagraph}
Massimiliano Mancini, Samuel~Rota Bulo, Barbara Caputo, and Elisa Ricci.
\newblock Adagraph: Unifying predictive and continuous domain adaptation
  through graphs.
\newblock In {\em Proceedings of the IEEE/CVF Conference on Computer Vision and
  Pattern Recognition}, pages 6568--6577, 2019.

\bibitem{masana2021ternary}
Marc Masana, Tinne Tuytelaars, and Joost van~de Weijer.
\newblock Ternary feature masks: zero-forgetting for task-incremental learning.
\newblock In {\em Proceedings of the IEEE/CVF Conference on Computer Vision and
  Pattern Recognition}, pages 3570--3579, 2021.

\bibitem{peng2017visda}
Xingchao Peng, Ben Usman, Neela Kaushik, Judy Hoffman, Dequan Wang, and Kate
  Saenko.
\newblock Visda: The visual domain adaptation challenge.
\newblock {\em arXiv preprint arXiv:1710.06924}, 2017.

\bibitem{saito2020universal}
Kuniaki Saito, Donghyun Kim, Stan Sclaroff, and Kate Saenko.
\newblock Universal domain adaptation through self supervision.
\newblock {\em Advances in Neural Information Processing Systems}, 33, 2020.

\bibitem{saito2017adversarial}
Kuniaki Saito, Yoshitaka Ushiku, Tatsuya Harada, and Kate Saenko.
\newblock Adversarial dropout regularization.
\newblock {\em ICLR}, 2018.

\bibitem{saito2018maximum}
Kuniaki Saito, Kohei Watanabe, Yoshitaka Ushiku, and Tatsuya Harada.
\newblock Maximum classifier discrepancy for unsupervised domain adaptation.
\newblock In {\em Proceedings of the IEEE Conference on Computer Vision and
  Pattern Recognition}, pages 3723--3732, 2018.

\bibitem{saito2018open}
Kuniaki Saito, Shohei Yamamoto, Yoshitaka Ushiku, and Tatsuya Harada.
\newblock Open set domain adaptation by backpropagation.
\newblock In {\em Proceedings of the European Conference on Computer Vision
  (ECCV)}, pages 153--168, 2018.

\bibitem{serra2018overcoming}
Joan Serra, Didac Suris, Marius Miron, and Alexandros Karatzoglou.
\newblock Overcoming catastrophic forgetting with hard attention to the task.
\newblock In {\em International Conference on Machine Learning}, pages
  4548--4557. PMLR, 2018.

\bibitem{shi2012information}
Yuan Shi and Fei Sha.
\newblock Information-theoretical learning of discriminative clusters for
  unsupervised domain adaptation.
\newblock In {\em Proceedings of the 29th International Coference on
  International Conference on Machine Learning}, pages 1275--1282, 2012.

\bibitem{shu2018dirt}
Rui Shu, Hung~H Bui, Hirokazu Narui, and Stefano Ermon.
\newblock A dirt-t approach to unsupervised domain adaptation.
\newblock {\em ICLR}, 2018.

\bibitem{su2020gradient}
Peng Su, Shixiang Tang, Peng Gao, Di Qiu, Ni Zhao, and Xiaogang Wang.
\newblock Gradient regularized contrastive learning for continual domain
  adaptation.
\newblock {\em AAAI}, 2021.

\bibitem{sun2016return}
Baochen Sun, Jiashi Feng, and Kate Saenko.
\newblock Return of frustratingly easy domain adaptation.
\newblock In {\em Thirtieth AAAI Conference on Artificial Intelligence}, 2016.

\bibitem{tang2020unsupervised}
Hui Tang, Ke Chen, and Kui Jia.
\newblock Unsupervised domain adaptation via structurally regularized deep
  clustering.
\newblock In {\em Proceedings of the IEEE/CVF Conference on Computer Vision and
  Pattern Recognition}, pages 8725--8735, 2020.

\bibitem{tzeng2014deep}
Eric Tzeng, Judy Hoffman, Ning Zhang, Kate Saenko, and Trevor Darrell.
\newblock Deep domain confusion: Maximizing for domain invariance.
\newblock {\em arXiv preprint arXiv:1412.3474}, 2014.

\bibitem{van2020scan}
Wouter Van~Gansbeke, Simon Vandenhende, Stamatios Georgoulis, Marc Proesmans,
  and Luc Van~Gool.
\newblock Scan: Learning to classify images without labels.
\newblock In {\em European Conference on Computer Vision}, pages 268--285.
  Springer, 2020.

\bibitem{venkateswara2017deep}
Hemanth Venkateswara, Jose Eusebio, Shayok Chakraborty, and Sethuraman
  Panchanathan.
\newblock Deep hashing network for unsupervised domain adaptation.
\newblock In {\em Proceedings of the IEEE Conference on Computer Vision and
  Pattern Recognition}, pages 5018--5027, 2017.

\bibitem{wu2020dual}
Yuan Wu, Diana Inkpen, and Ahmed El-Roby.
\newblock Dual mixup regularized learning for adversarial domain adaptation.
\newblock {\em ECCV}, 2020.

\bibitem{wu2018unsupervised}
Zhirong Wu, Yuanjun Xiong, Stella~X Yu, and Dahua Lin.
\newblock Unsupervised feature learning via non-parametric instance
  discrimination.
\newblock In {\em Proceedings of the IEEE Conference on Computer Vision and
  Pattern Recognition}, pages 3733--3742, 2018.

\bibitem{Xu_2019_ICCV}
Ruijia Xu, Guanbin Li, Jihan Yang, and Liang Lin.
\newblock Larger norm more transferable: An adaptive feature norm approach for
  unsupervised domain adaptation.
\newblock In {\em The IEEE International Conference on Computer Vision (ICCV)},
  October 2019.

\bibitem{yang2020bi}
Guanglei Yang, Haifeng Xia, Mingli Ding, and Zhengming Ding.
\newblock Bi-directional generation for unsupervised domain adaptation.
\newblock In {\em AAAI}, pages 6615--6622, 2020.

\bibitem{yang2020unsupervised}
Shiqi Yang, Yaxing Wang, Joost van~de Weijer, Luis Herranz, and Shangling Jui.
\newblock Unsupervised domain adaptation without source data by casting a bait.
\newblock {\em arXiv preprint arXiv:2010.12427}, 2020.

\bibitem{ye2020light}
Shaokai Ye, Kailu Wu, Mu Zhou, Yunfei Yang, Sia~Huat Tan, Kaidi Xu, Jiebo Song,
  Chenglong Bao, and Kaisheng Ma.
\newblock Light-weight calibrator: a separable component for unsupervised
  domain adaptation.
\newblock In {\em Proceedings of the IEEE/CVF conference on computer vision and
  pattern recognition}, pages 13736--13745, 2020.

\bibitem{you2019universal}
Kaichao You, Mingsheng Long, Zhangjie Cao, Jianmin Wang, and Michael~I Jordan.
\newblock Universal domain adaptation.
\newblock In {\em Proceedings of the IEEE Conference on Computer Vision and
  Pattern Recognition}, pages 2720--2729, 2019.

\bibitem{zhang2019bridging}
Yuchen Zhang, Tianle Liu, Mingsheng Long, and Michael Jordan.
\newblock Bridging theory and algorithm for domain adaptation.
\newblock In {\em International Conference on Machine Learning}, pages
  7404--7413, 2019.

\bibitem{zhuang2019local}
Chengxu Zhuang, Alex~Lin Zhai, and Daniel Yamins.
\newblock Local aggregation for unsupervised learning of visual embeddings.
\newblock In {\em Proceedings of the IEEE/CVF International Conference on
  Computer Vision}, pages 6002--6012, 2019.

\end{thebibliography}
}

\end{document}